\documentclass[letterpaper, 10 pt, conference]{ieeeconf}
\IEEEoverridecommandlockouts
\usepackage{cite}
\usepackage{amsmath,amssymb,amsfonts}
\usepackage{algorithm}
\usepackage{algpseudocode} 
\usepackage{float}         
\usepackage{stfloats} 
\usepackage{graphicx}
\usepackage{textcomp}
\usepackage{xcolor}
\usepackage{lipsum}
\usepackage{soul, color}
\def\BibTeX{{\rm B\kern-.05em{\sc i\kern-.025em b}\kern-.08em
    T\kern-.1667em\lower.7ex\hbox{E}\kern-.125emX}}
    
\title{\LARGE \bf 
Monolithic Units: Actuation, Sensing, and Simulation for Integrated Soft Robot Design}

\author{Trevor Exley$^{1}$, Anderson Brazil Nardin$^{1}$, Petr Trunin$^{1,2}$, Diana Cafiso$^{1}$, Lucia Beccai$^{1,\dagger}$ %
\thanks{This work was funded by the European Union Horizon 2020 research and innovation programme under grant agreement No. 863212 (PROBOSCIS project).}%
\thanks{$^{1}$T. Exley, A.B. Nardin, P. Trunin, D. Cafiso, and L. Beccai are with Soft BioRobotics Perception Lab of the Istituto Italiano di Tecnologia (IIT), Genova 16163, Italy (lucia.beccai@iit.it)%
}%
\thanks{$^{2}$P. Trunin is with the Open University ARC@IIT, Istituto Italiano di Tecnologia, 16163, Genova, Italy.}
}
\begin{document}

\maketitle

\begin{abstract}
This work introduces the Monolithic Unit (MU), an actuator–lattice–sensor building block for soft robotics. The MU integrates pneumatic actuation, a compliant lattice envelope, and candidate sites for optical waveguide sensing into a single printed body. In order to study reproducibility and scalability, a parametric design framework establishes deterministic rules linking actuator chamber dimensions to lattice unit cell size. Experimental homogenization of lattice specimens provides effective material properties for finite element simulation. Within this simulation environment, sensor placement is treated as a discrete optimization problem, where a finite set of candidate waveguide paths derived from lattice nodes is evaluated by introducing local stiffening, and the configuration minimizing deviation from baseline mechanical response is selected. Optimized models are fabricated and experimentally characterized, validating the preservation of mechanical performance while enabling embedded sensing. The workflow is further extended to scaled units and a two-finger gripper, demonstrating generality of the MU concept. This approach advances monolithic soft robotic design by combining reproducible co-design rules with simulation-informed sensor integration.
\end{abstract}


\section{Introduction}

Monolithic soft robotic devices aim to integrate actuation and sensing within a single printed body, avoiding multi-material interfaces, manual assembly, and post-processing that degrade compliance and repeatability. Additive manufacturing enables increasingly complex internal features, yet realizing monolithic devices that maintain softness, airtightness, and functional sensing remains challenging due to process limits, leakage, and stiffness mismatches introduced by heterogeneous elements.  

Historically, soft robotic bodies are fabricated by molding elastomers or assembling fabric and pouch structures, which require multiple materials and manual steps that complicate integration of internal features. Recent advances in additive manufacturing enable single-process builds with sealed cavities, architected lattices, and embedded sensing, reducing assembly and improving reproducibility \cite{zhaiDesktopFabricationMonolithic2023,tawkFully3DPrinted2019,truninMELEGROSMonolithicElephantinspired2025}. This shift motivates the development of standardized monolithic building blocks that couple actuation, structure, and sensing within one printed body.

Architected lattices provide a suitable substrate for monolithic integration. Triply periodic minimal surface (TPMS) and related cellular architectures offer tunable stiffness, strength, and anisotropy, distribute loads efficiently, and facilitate support-lean printing of internal cavities \cite{altamimiStiffnessStrengthAnisotropy2025,vietMechanicalAttributesWave2022,xiaoLargeDeformationResponse2024a,najiHybridPlateTPMSLattice2025,novakDevelopmentNovelHybrid2021,zhaoTPMSbasedInterpenetratingLattice2023}. In practice, CAD and AM toolchains play a central role in controlling cell topology, unit-cell size, and strut or ligament morphology, which directly affects printability and mechanical response \cite{geyerComparisonCADSoftware2024}.  

Recent demonstrations of monolithic devices have shown that functional actuation and sensing can be achieved in a single print \cite{zhaiDesktopFabricationMonolithic2023,tawkFully3DPrinted2019,yaoJAMMitMonolithic3DPrinting2025,truninMELEGROSMonolithicElephantinspired2025,kiSoftGripperMovable2025a,wuMonolithicProgrammableFabricStacking2025}. These examples highlight the feasibility of monolithic systems, but sensor placement is typically determined heuristically and design iterations are required to compensate for mechanical changes introduced by embedded features. Other monolithic or hybrid designs have emphasized embedded sensing or variable stiffness mechanisms \cite{kiSoftGripperMovable2025a}, or new programmable fabrication strategies \cite{wuMonolithicProgrammableFabricStacking2025}, yet still rely on trial-and-error approaches rather than simulation-informed integration.  

Simulation in soft robotics has been more frequently applied to actuator design or sensor optimization. Examples include automated sensor placement frameworks such as MakeSense \cite{tapiaMakeSenseAutomatedSensor2020}, co-learning approaches that optimize both task and sensor location \cite{spielbergCoLearningTaskSensor2021}, differentiable simulation for pneumatic actuator geometry \cite{gjokaSoftPneumaticActuator2024}, and multifunctional graded lattice systems with integrated sensing \cite{yangInterpenetratingPhaseComposite2023}. Broader reviews of modeling techniques emphasize the challenges of nonlinear deformation, large-strain mechanics, and fluid–structure coupling, which limit predictive accuracy \cite{qinModelingSimulationDynamics2023}. These methods improve actuation or sensing performance but rarely quantify or minimize the perturbation to mechanical response that occurs when embedding sensing elements into structured lattices.  

Soft robotics sensing more broadly has been approached through separate sensing \cite{shihElectronicSkinsMachine2020a,wangSensingExpectationEnables2024,xuOpticalLaceSynthetic2019}, resistive transduction \cite{liuTouchlessInteractiveTeaching2022,cafisoDLPPrintablePorousCryogels2024,soShapeEstimationSoft2021,jiDesignCalibration3D2023}, and capacitive transduction \cite{kimInherentlyIntegratedMicrofiberbased2024,huStretchableEskinTransformer2023,hashizumeCapacitiveSensingGripper2019,niuHighlyMorphologyControllableHighly2020}. These approaches expand the sensing repertoire but introduce mismatched mechanical properties, reducing durability and complicating integration in fully monolithic designs. Optical waveguides have recently been introduced as an alternative, fabricated directly by stereolithography, with geometry constrained by minimum thickness and surface pattern requirements \cite{truninDesign3DPrinting2025}.  

\begin{figure*}[t!]
    \centering
    \includegraphics[width=\textwidth]{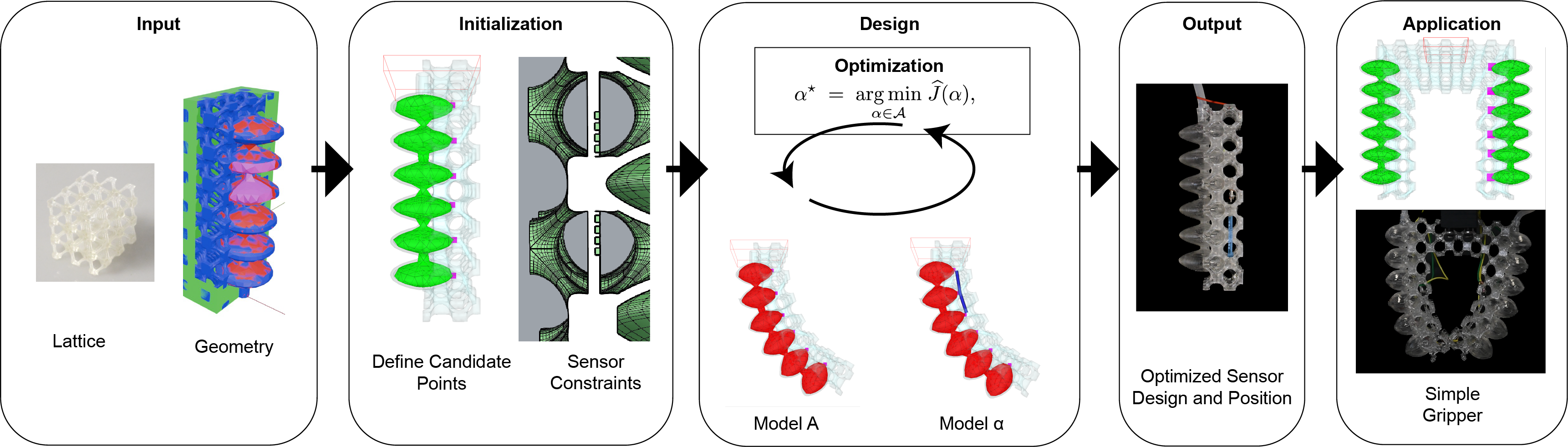}
    \caption{Overall workflow for the Monolithic Unit (MU). 
    Input: a lattice unit cell and actuator geometry define the base structure. 
    Initialization: candidate sensor points are extracted from lattice nodes and constrained by manufacturable sensor geometry. 
    Design: simulation of the baseline (Model~A) and sensorized (Model~$\alpha$) configurations identifies the path minimizing curvature deviation. 
    Output: printed MU incorporating the selected sensor layout. 
    Application: demonstration of the sensorized MU integrated into a simple gripper.}
    \label{fig:design}
\end{figure*}

This study introduces a Monolithic Unit (MU) paradigm that establishes an actuator–lattice block which integrates optical waveguide sensors. The workflow begins with experimental homogenization of lattice stiffness, followed by simulation of the actuator–lattice body without sensors to establish baseline kinematics (Model A). Candidate waveguide paths are generated from lattice node coordinates and evaluated as locally stiffened inclusions (Model $\alpha$). The selection criterion minimizes deviation of nodal displacement norms from the baseline, thereby preserving actuation response. The optimal configuration is then printed and characterized. Optical waveguide constraints inform the lattice strut thickness at the reference scale and illustrate how sensing requirements couple to lattice geometry. Scalability is evaluated at $0.75\times$, $1.00\times$, and $1.50\times$, and a grasping application is demonstrated by extending the same MU-based process to a two-finger gripper.  

The contributions of this work are: (i) a reproducible MU design rule set that relates actuator geometry to lattice topology, (ii) a simulation-informed method for sensor integration that minimizes mechanical deviation relative to a baseline design, and (iii) experimental validation across scales and in a two-finger configuration. The overall workflow and MU concept are summarized in Figure \ref{fig:design}.

\section{Monolithic Unit}

The Monolithic Unit (MU) is introduced as an actuator–lattice–sensor building block for soft robotic systems. It combines a pneumatic actuator, a compliant lattice envelope, and an integrated optical waveguide within a single printed body. The MU is designed to be reproducible, scalable, and suitable as a foundation for simulation-based evaluation of sensor integration. 

\begin{figure}[htb!]
    \centering
    \includegraphics[width=\linewidth]{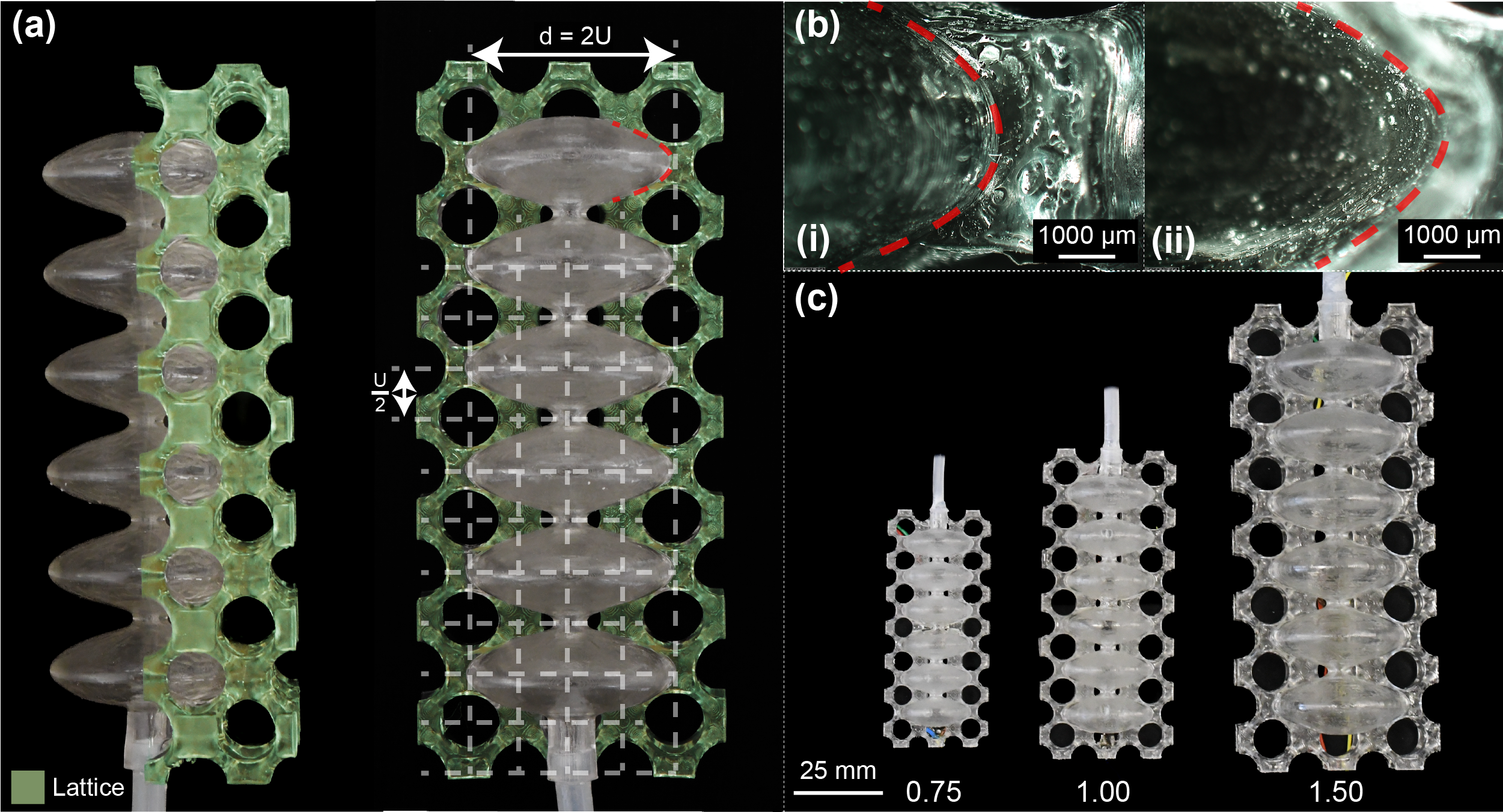}
    \caption{Design and fabrication of the Monolithic Unit. 
    (a) Co-designed actuator and lattice geometry showing the body-centered lattice (green) and pneumatic cavities (gray). 
    (b) Microscopy images of internal lattice and cross-section of the membrane junctions (outlined in red) confirming print fidelity. 
    (c) Fabricated MUs at scales $0.75\times$, $1.00\times$, and $1.50\times$}
    \label{fig:codesign}
\end{figure}

The actuator consists of six bladder chambers arranged serially along a central axis. Each chamber includes a bladder and neck region defined by parabolic profiles revolved around the axis to ensure curvature continuity and reduce stress concentrations during cyclic pressurization. Chamber spacing equals the lattice unit cell size $U$, and the bladder diameter $d$ is set to $2U$, linking actuator geometry directly to lattice parameters (Figure~\ref{fig:codesign}a).

The actuator and lattice are co-designed so that lattice nodes coincide with the apices of each bladder, creating tangential anchoring that naturally limits radial expansion while promoting axial deformation. This co-design supports smooth load transfer between membrane and lattice and stabilizes thin features during fabrication. The high-resolution SLA process ensures clean integration between the two domains, as observed in microscopy (Figure~\ref{fig:codesign}b(i)), while the interior surfaces of the bladder remain uniform without material pooling (Figure~\ref{fig:codesign}b(ii)).

The MU also defines a candidate space for optical waveguide integration. Node coordinates are extracted from the lattice geometry, and those approximately aligned with the bending plane are selected as potential sensing paths for simulation-based evaluation. Each waveguide is a transparent channel integrated into the lattice structure and patterned with a series of shallow surface wells that enhance sensitivity to bending. A light source is placed at one end of the channel, while a photodetector positioned at the opposite end records transmitted intensity. During deformation, bending of the waveguide induces localized scattering at the patterned surface, resulting in a measurable reduction of detected light. The periodic wells, $1.0$ mm wide and $0.65$ mm deep, are directly printed on the outer surface of the waveguide, ensuring repeatable fabrication. This configuration allows simulation-based assessment of how embedded sensing paths influence actuator mechanics and provides a basis for subsequent sensor placement optimization.

All geometric rules of the Monolithic Unit (MU) scale with the lattice unit cell size $U$. The $1.0\times$ MU is defined by $U = 12.5,\mathrm{mm}$ and a minimum strut thickness of $1.5,\mathrm{mm}$, determined by the waveguide dimensions and printer resolution. Scaled variants ($0.75\times$ and $1.5\times$) preserve these proportional relationships of chamber spacing, bladder diameter, and node–apex alignment, thereby maintaining consistent strain-limiting behavior across sizes (Figure~\ref{fig:codesign}c). The only feature that changes with scale is the length of the integrated waveguides, since the spacing between lattice nodes serving as anchor points varies with $U$.
Consequently, the waveguide patterning (i.e., the number of surface wells) adjusts proportionally to accommodate the different lengths while preserving the same surface well geometry and periodicity.

\section{Materials and Methods}

\subsection{Fabrication}
All parts were printed in Elastic~50A Resin (Formlabs, USA) on a Form~4 stereolithography (SLA) printer at 0.1 mm layer height. Prints followed the manufacturer workflow: perimeter/model/support regions at 38.40 mJ/cm\textsuperscript{2} with 11.5 mW/cm\textsuperscript{2} light intensity. Post-processing consisted of a 20 min isopropyl alcohol wash (Form Wash) and a 30 min, 70$^\circ$C UV cure (Form Cure).

\subsection{Parametric Design of the Monolithic Unit}
The MU was parametrically defined in Grasshopper (Rhinoceros~3D). The actuator comprises six chambers generated by revolving parabolic bladder/neck profiles about the central axis. Chamber spacing equals the lattice unit cell size $U$ and the bladder diameter $d=2U$ at the reference scale $U = 12.5$ mm. From this model, the following files are exported: (i) STEP files for cavities, membranes, and lattice envelope (deformation/collision), (ii) an STL of the full MU for printing/visualization, and (iii) a TXT file with lattice node coordinates to define candidate sensor paths.

\subsection{Lattice Generation}
A body-centered cell is instantiated in a parallelepiped envelope around the actuator using the \textit{Crystallon} plugin for Grasshopper. Struts are thickened via the \textit{MultiPipe} component with a taper that narrows near junctions and widens mid-span, reducing bulky nodes, improving printability, and qualitatively emulating ligament-like IWP morphology while reducing file complexity \cite{liuMemoryEfficientModelingSlicing2021,mirzavandInvestigationMechanicalCharacteristics2025}. The minimum strut thickness at $1.0\times$ is 1.5 mm, set by optical waveguide patterns. Compression tests on cubic lattice specimens ($N {=} 3$ per scale) provide effective moduli for homogenized simulation: $21.70 \pm 0.35$ kPa ($0.75\times$), $18.34 \pm 1.86$ kPa ($1.00\times$), and $16.38 \pm 1.21$ kPa ($1.50\times$).

\subsection{Simulation Workflow}

To achieve this objective, and building on recent results in simulating soft lattice structures \cite{nardin_exploring_2025}, the proposed method is based on a workflow linking design and fabrication through simulation in SOFA (Simulation Open Framework Architecture)~\cite{duriez_realistic_2006, payan_sofa_2012} to design and position the soft optical sensors. All parts were meshed in Gmsh~\cite{geuzaine_gmsh_2009} through a Python interface, ensuring compatibility between the geometric models used for visualization, deformation, and collision handling. 

\begin{figure}[htb!]
    \centering
    \includegraphics[width=\linewidth]{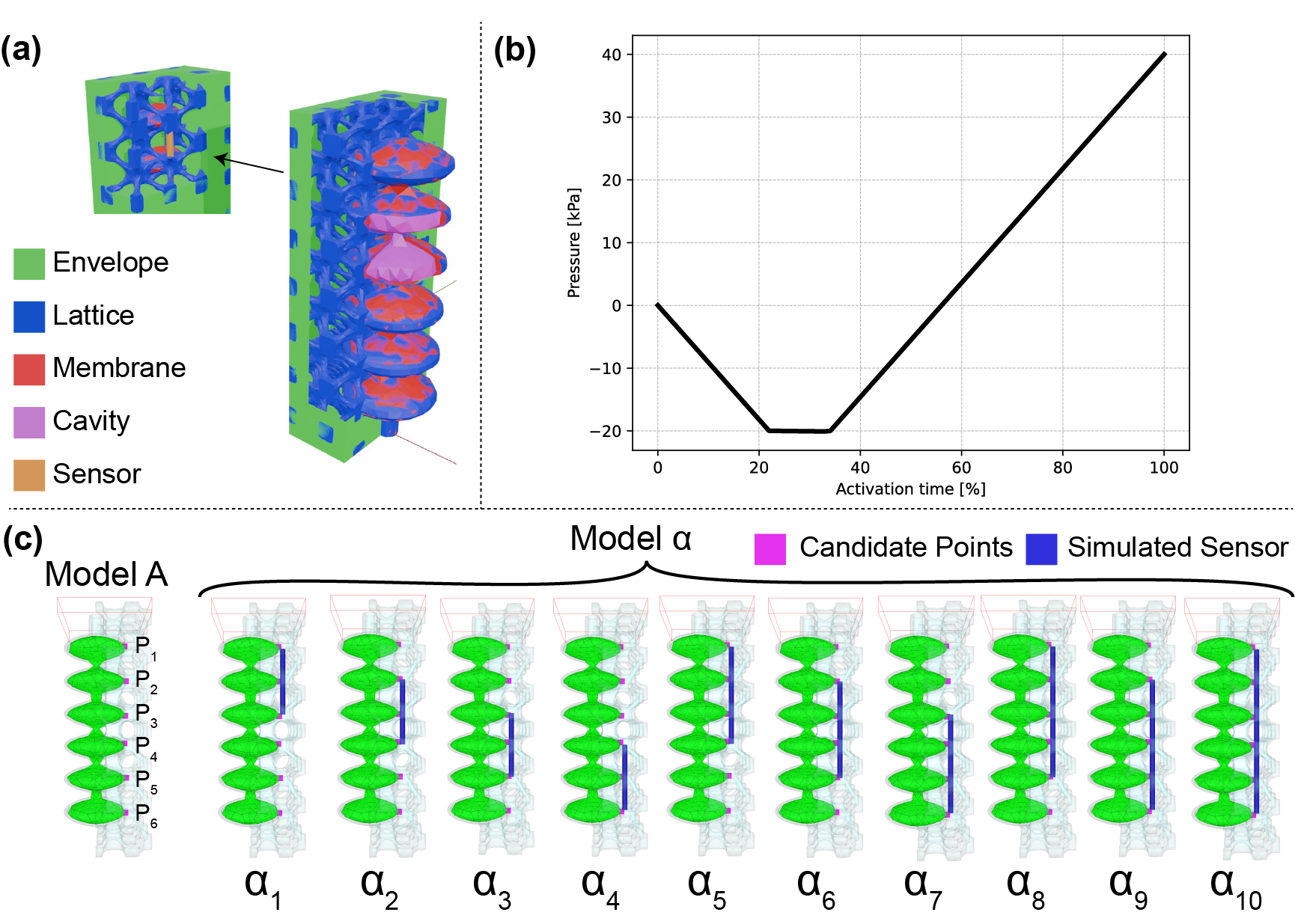}
    \caption{Simulation of the MU. 
    (a) Broken-out view of components used in simulation: envelope, lattice, membrane, cavity, and sensor domains. 
    (b) Pressure actuation profile used to drive deformation during bending cycles. 
    (c) Model~A (no sensors) and Models~$\alpha_1$–$\alpha_{10}$ (sensorized configurations) showing different simulated sensor lengths and positions.}
    \label{fig:simulation}
\end{figure}

The simulation environment reconstructs the MU from three geometric components: the actuator cavities, the membrane regions, and the homogenized lattice envelope (Figure~\ref{fig:simulation}a). The cavity surfaces define the pneumatic loading boundaries, while the membranes and lattice represent regions of higher and lower stiffness, respectively. The meshed bodies are imported into SOFA, where each element is assigned its corresponding material properties, experimentally determined for the lattice and nominal for the membranes, and assembled into a single deformable model governed by large-deformation tetrahedral finite elements.

Pneumatic inputs are prescribed as a time-varying pressure field applied to the inner cavity surfaces. The pressure evolution follows a predefined actuation protocol stored in a CSV file, corresponding to the experimental sequence used for a simple grasp: starting from atmospheric conditions, pressure decreases to $-20 \mathrm{kPa}$ to produce extension and subsequently rises to $40 \mathrm{kPa}$ to achieve flexion (Figure~\ref{fig:simulation}b).

During each simulation, specific lattice-core nodes along the sagittal plane are continuously monitored to reconstruct the actuator backbone and evaluate its kinematic behavior. These points are selected from the cores of the lattice unit cells, which coincide with potential sensor anchoring sites. The monitored positions are exported at each time step for post-processing and deviation analysis relative to the baseline actuator.

To evaluate multiple sensor configurations, a single validated baseline mesh is reused, and local material properties are modified only in the regions corresponding to virtual sensor paths. Each simulation variant defines one candidate model, in which the stiffened inclusion replaces a portion of the homogenized lattice while maintaining identical loads and boundary conditions. 

The comparison across candidates is performed through an exhaustive search, a solving approach that systematically evaluates every possible configuration to identify the optimal solution \cite{goos_exhaustive_2000, taimoor_holistic_2022}. This approach guarantees convergence to the global optimum when the search space is finite and discretized, making it appropriate for the limited number of sensor placements considered in this work. All simulations are executed in parallel, enabling simultaneous evaluation of multiple configurations while preserving strict comparability between results. This approach ensures that the deviations observed in monitored trajectories isolate only the mechanical influence of each virtual sensor configuration.

\subsection{Optimization of Sensor Placement}
\label{subsec:optimization}

The objective is to embed an optical waveguide path that preserves the native mechanics of the actuator-lattice body. Let Model~A denote the baseline (no sensors) under a prescribed actuation program, and let Model~$\alpha$ denote a candidate sensorized variant obtained by introducing a locally stiffened inclusion along a path defined on the sagittal plane.  

Candidate paths are constructed from contiguous sequences of anchor nodes located at the cores of the lattice unit cells that intersect the sagittal plane of the actuator. 
Let $\mathcal{P} = \{\, P_1, P_2, \ldots, P_n \,\}$ be the points which represent the ordered set of these $n$ available anchors, where $P_i$ is the $i$-th node along the sagittal direction (Figure~\ref{fig:simulation}c). 
Let $i, h \in \mathbb{Z}^{+}$ denote integer indices representing, respectively, the starting position and the length of a contiguous subsequence, with 
$3 \le h \le n$ and $1 \le i \le n-h+1$. 
Each contiguous group of length $h$ starting at index $i$ is obtained through a \emph{sliding-window} (ordered sequence) operator:
\begin{equation}
  G_{i,h}
  \;=\;
  [\, P_i,\, P_{i+1},\, P_{i+2},\, \ldots,\, P_{i+h-1} \,].
  \label{eq:slidingwindow}
\end{equation}
The complete set of all admissible contiguous groups (graphically represented in Figure~\ref{fig:simulation}c) is therefore
\begin{equation}
\begin{aligned}
  \mathcal{G}
  \;&=\;
  \{\, G_{i,h} \;\big|\; h \in [3,n],\; i \in [1,n-h+1] \,\} \\[4pt]
  \;&=\;
  \bigcup_{h=3}^{n}
  \;\bigcup_{i=1}^{\,n-h+1}
  \{\, [\, P_i,\, \ldots,\, P_{i+h-1} \,] \,\}.
\end{aligned}
\label{eq:allcandidates}
\end{equation}

Here $G_{i,h}$ denotes a contiguous subsequence (window) of anchors, 
and $\mathcal{G}$ collects the union of all possible windows satisfying the contiguity constraint $h \ge 3$. 
Each element of $\mathcal{G}$ defines one candidate path for virtual sensor placement.

For each candidate, the physical sensor is represented virtually by a cylindrical inclusion of radius $r_s$ ($1\,\mathrm{mm}$) aligned with the anchor sequence; on the unmodified baseline tetrahedral mesh, a region-of-interest (ROI) is computed by intersecting the inclusion with the mesh and assigning a bulk-like Young’s modulus $E_{\mathrm{sens}}$ to the intersected tetrahedra, while all remaining lattice elements retain the homogenized modulus $E_{\mathrm{lat}}$. 
Let $\Omega\subset\mathbb{R}^3$ denote the solid domain of the model and let $\Omega_{\mathrm{ROI}}(\alpha)\subset\Omega$ be the inclusion volume associated with candidate $\alpha$. 
A spatial point is written as $\mathbf{x}=(x_1,x_2,x_3)\in\Omega$, and the Young’s modulus is a spatially varying scalar field $E:\Omega\to\mathbb{R}^{+}$. 
The set difference used is written as $\Omega\setminus\Omega_{\mathrm{ROI}}(\alpha)=\{\mathbf{x}\in\Omega \mid \mathbf{x}\notin\Omega_{\mathrm{ROI}}(\alpha)\}$. 
With these definitions, the spatial distribution of stiffness is
\begin{equation}
E(\mathbf{x}) \;=\; 
\begin{cases}
E_{\mathrm{sens}}, & \mathbf{x} \in \Omega_{\mathrm{ROI}}(\alpha),\\[2pt]
E_{\mathrm{lat}},  & \mathbf{x} \in \Omega\setminus \Omega_{\mathrm{ROI}}(\alpha),
\end{cases}
\qquad E_{\mathrm{sens}} \gg E_{\mathrm{lat}} ,
\label{eq:piecewiseE}
\end{equation}
with geometry, boundary conditions, and actuation loads identical to Model~A.

The analysis compares Model~A and Model~$\alpha$ using the same monitored nodes extracted from the unit–cell cores that lie on the actuator’s sagittal plane (i.e., the cores of the lattice cells intersecting that plane). Let $\mathbf{X}(t)\in\mathbb{R}^{N\times3}$ represent the ordered three-dimensional coordinates of the monitored polyline at time $t$. To achieve consistent spatial correspondence, each polyline is reparameterized by a normalized arc-length parameter $\ell$. Let $\mathbf{r}_A(t,\ell)$ and $\mathbf{r}_\alpha(t,\ell)$ be the resulting continuous centerlines for Model~A and Model~$\alpha$, respectively, obtained by cubic spline interpolation of the discrete monitored nodes. The instantaneous deviation at a given actuation time and arc-length is defined as
\begin{equation}
\Delta(t,\ell)\;=\;\bigl\|\,\mathbf{r}_\alpha(t,\ell)\;-\;\mathbf{r}_A(t,\ell)\,\bigr\|_2 .
\label{eq:delta}
\end{equation}

In practice, the actuation interval is sampled uniformly at \(k\) normalized time instants \(\{t_m\}_{m=1}^{k}\), 
and the arc-length is discretized uniformly at \(j\) locations \(\{\ell_s\}_{s=1}^{j}\), defined as
\begin{equation}
t_m = \frac{m-1}{k-1}, \quad 
\ell_s = \frac{s-1}{j-1}, \qquad 
t_m,\,\ell_s \in [0,1].
\label{eq:sampling}
\end{equation}

The resulting deviation matrix has entries
\begin{equation}
\Delta_{m s}\;=\;\Delta\!\left(t_m,\ell_s\right), 
\qquad m=1,\dots,k,\;\; s=1,\dots,j .
\label{eq:delta_matrix}
\end{equation}
Two marginal summaries are useful for interpretation: the length-averaged deviation over time
\begin{equation}
\overline{\Delta}(t_m)\;=\;\frac{1}{j}\sum_{s=1}^{j} \Delta_{m s},
\label{eq:avg_over_length}
\end{equation}
and the time-averaged deviation over length
\begin{equation}
\widetilde{\Delta}(\ell_s)\;=\;\frac{1}{k}\sum_{m=1}^{k} \Delta_{m s}.
\label{eq:avg_over_time}
\end{equation}
The scalar objective used for exhaustive search is the global average deviation,
\begin{equation}
\widehat{J}(\alpha)\;=\;\frac{1}{k\,j}\sum_{m=1}^{k}\sum_{s=1}^{j}\Delta_{m s},
\label{eq:objective}
\end{equation}
which is unweighted and directly reflects the aggregate geometric discrepancy between Model~\(\alpha\) and Model~A over the actuation cycle and along the actuator length. The optimal placement is selected by
\begin{equation}
\alpha^\star \;=\;\underset{\alpha\in\mathcal{A}}{\arg\min}\;\widehat{J}(\alpha),
\label{eq:argmin}
\end{equation}
where \(\mathcal{A}\) is the finite set of all contiguous anchor sequences considered.

While the deviation metric in~\eqref{eq:objective} is evaluated discretely over sampled times and arc-length locations, a continuous formulation can be introduced to emphasize its conceptual meaning as an $L^2$-type norm over both the temporal and spatial domains. 
Let \(T\) denote the actuation duration and \(L\) the actuator length along the sagittal direction. 
The continuous nodewise discrepancy functional is then defined as
\begin{equation}
J_{\mathrm{cont}}(\alpha)
\;=\;
\left(
\frac{1}{T\,L}
\int_{0}^{T}
\int_{0}^{L}
\bigl\|
\mathbf{r}_\alpha(t,\ell)
-
\mathbf{r}_A(t,\ell)
\bigr\|_2^{2}
\, d\ell\, dt
\right)^{\!1/2}.
\label{eq:Jcont}
\end{equation}
This expression represents the theoretical continuous counterpart of the discrete deviation measures employed in practice, capturing the root-mean-square geometric discrepancy between the two models over the entire spatiotemporal domain.

\begin{algorithm}[!htbp]
  \caption{Virtual-Stiffening for Discrete Optimization of Sensor Placement on a Single Baseline Mesh}
  \label{alg:optimization}
  \begin{algorithmic}[1]
    \State Simulate \textbf{Model~A} on the baseline mesh with homogenized lattice modulus \(E_{\mathrm{lat}}\); record monitored node trajectories \(\mathbf{X}_A(t)\) over the actuation interval
    \State Choose \(k\) uniformly spaced normalized time samples \(t_m\); construct the arc-length parametrization \(\mathbf{r}_A(t,\ell)\) by cubic splines on the monitored nodes
    \For{\textbf{each} candidate path \(\alpha \in \mathcal{A}\)}
        \State Build the virtual inclusion as a union of cylinders of radius \(r_s\) along the anchor sequence; compute the mesh ROI by intersection with the baseline tetrahedra
        \State Assign \(E(\mathbf{x})\) as in~\eqref{eq:piecewiseE}; keep all loads and boundary conditions identical to Model~A; simulate \textbf{Model~\(\alpha\)} and record \(\mathbf{X}_\alpha(t)\)
        \State Consider \(t_m\), reparameterize the monitored polyline to obtain \(\mathbf{r}_\alpha(t,\ell)\) on the same arc-length grid as \(\mathbf{r}_A(t,\ell)\); assemble \(\Delta_{m s}\) via~\eqref{eq:delta}--\eqref{eq:delta_matrix}
        \State Compute \(\widehat{J}(\alpha)\) from~\eqref{eq:objective}
    \EndFor
    \State Select \(\alpha^\star\) according to~\eqref{eq:argmin}; report \(\Delta_{m s}\) as a heatmap (length\% \(\times\) time\%), the summaries~\eqref{eq:avg_over_length}--\eqref{eq:avg_over_time}, and \(\widehat{J}(\alpha)\) for ranking
  \end{algorithmic}
\end{algorithm}

The optimization workflow summarized in Algorithm~\ref{alg:optimization} 
reuses a single mesh (from Model~A) and solver configuration, modifying only the local material field in the sensor ROIs (to Model~\(\alpha\)). 
Because each candidate is evaluated under identical numerical and loading conditions, 
differences in $\widehat{J}(\alpha)$ isolate the mechanical effect of the candidate sensor path. 
The representation in~\eqref{eq:piecewiseE} captures the dominant mechanism by which an embedded waveguide perturbs the lattice domain, 
while the spline-based normalization ensures a valid geometric comparison in both time and length for all simulated configurations.

\subsection{Experimental Characterization}
Cubic lattice specimens at $0.75\times$, $1.0\times$, and $1.5\times$ scales ($N{=}3$ per scale) were tested in uniaxial compression to 40\% strain at 10 mm/min to obtain the effective moduli listed above. MU bending was evaluated under pneumatic inputs from $-20$ to $50$ kPa. Bending angles were extracted from video using ImageJ~\cite{schneiderNIHImageImageJ2012}.

For sensorized configurations, optical signals were acquired using infrared emitters (VSMY1850, 850 nm peak) and photoreceivers (VEMT7100X01; both Vishay Semiconductors) on a custom PCB. Data acquisition used Python~3.13; analysis was performed in MATLAB (The MathWorks, Inc.).

\begin{figure*}[ht!]
    \centering
    \includegraphics[width=\textwidth]{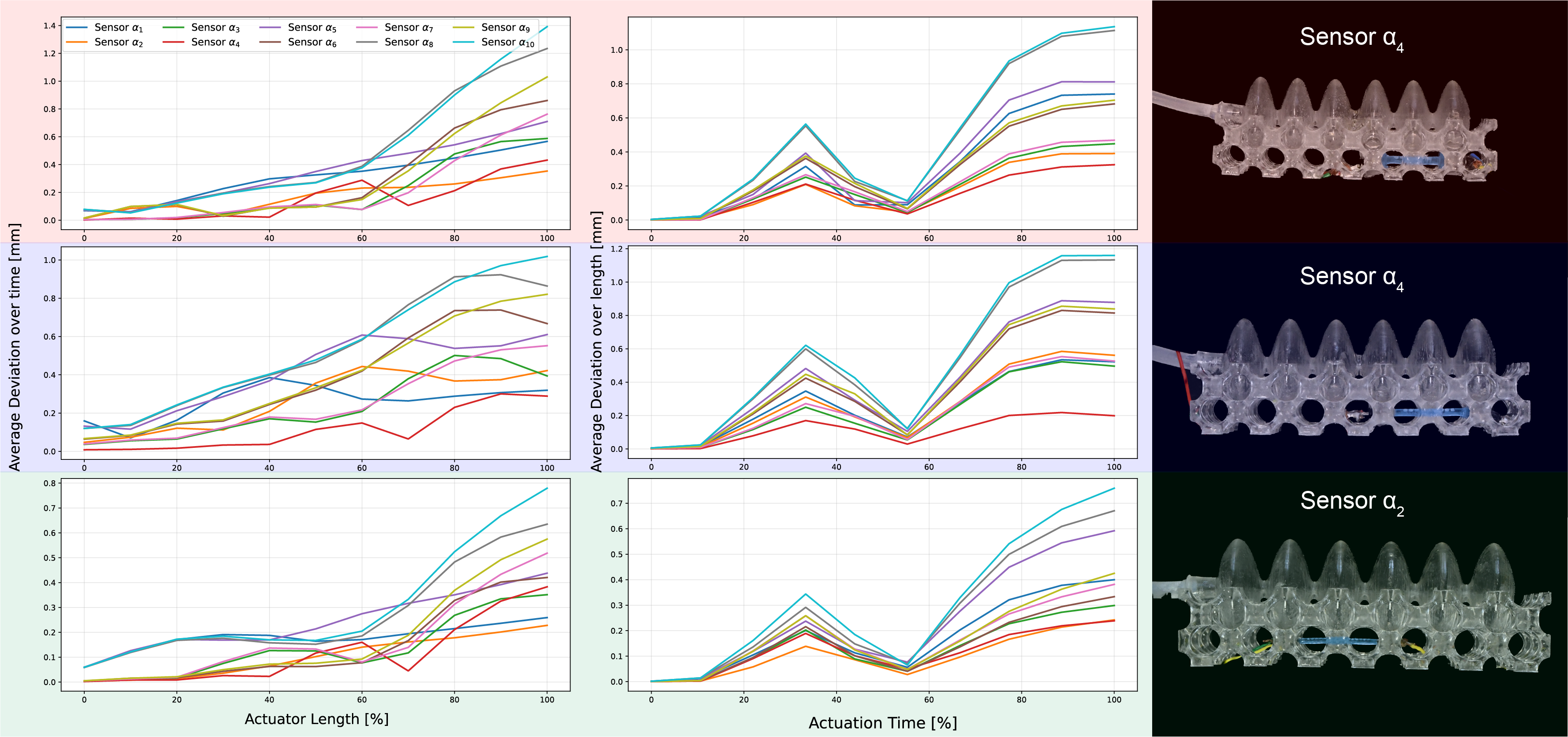}
    \caption{Evaluation of simulated sensing configurations. 
    Average deviation along actuator length (left) and actuation time (middle) for each sensor configuration $\alpha$ at three scales: $0.75\times$ (red), $1.0\times$ (blue), and $1.5\times$ (green). 
    Fabricated models (right) highlight sensors $\alpha_4$ and $\alpha_2$, which exhibit minimal deviation from the baseline Model~A.}
    \label{fig:optimization}
\end{figure*}

\section{Results}
\subsection{Simulation and Sensor Path Evaluation}
The simulation framework was used to compare the baseline configuration (Model~A) with ten candidate sensorized configurations (Models~$\alpha_1$–$\alpha_{10}$). Each configuration corresponds to a distinct continuous path of variable length along the actuator’s sagittal plane.  
Average deviation between Models~A and~$\alpha$ was evaluated over both actuator length and actuation time for three scales: $0.75\times$ (red), $1.0\times$ (blue), and $1.5\times$ (green) (Figure~\ref{fig:optimization}). Accordingly, sensor placement is reported as the discrete optimum $\alpha^\star$ obtained by the exhaustive virtual-stiffening over $\mathcal{A}$.

\begin{table}[ht!]
\centering
\caption{Global average deviation $\widehat{J}(\alpha_i)$ [mm] for each sensor $\alpha_i$ across different scales. Lower values indicate better performance.}
\begin{tabular}{c|cccc}
\hline
\textbf{Sensor} & \textbf{0.75$\times$} & \textbf{1.00$\times$} & \textbf{1.50$\times$} & \textbf{Gripper} \\ 
\hline
$\alpha_1$ & 0.308 & 0.248 & 0.179 & 0.262 \\
$\alpha_2$ & 0.176 & 0.376 & \textbf{0.103} & 0.268 \\
$\alpha_3$ & 0.203 & 0.315 & 0.137 & 0.233 \\
$\alpha_4$ & \textbf{0.153} & \textbf{0.155} & 0.119 & \textbf{0.114} \\
$\alpha_5$ & 0.351 & 0.436 & 0.244 & 0.411 \\
$\alpha_6$ & 0.302 & 0.493 & 0.145 & 0.379 \\
$\alpha_7$ & 0.216 & 0.351 & 0.170 & 0.251 \\
$\alpha_8$ & 0.479 & 0.550 & 0.275 & 0.522 \\
$\alpha_9$ & 0.313 & 0.522 & 0.178 & 0.395 \\
$\alpha_{10}$ & 0.490 & 0.575 & 0.308 & 0.538 \\
\hline
\end{tabular}
\label{tab:sensor_deviation}
\end{table}

\begin{figure}[htb!]
    \centering
    \includegraphics[width=.9\linewidth]{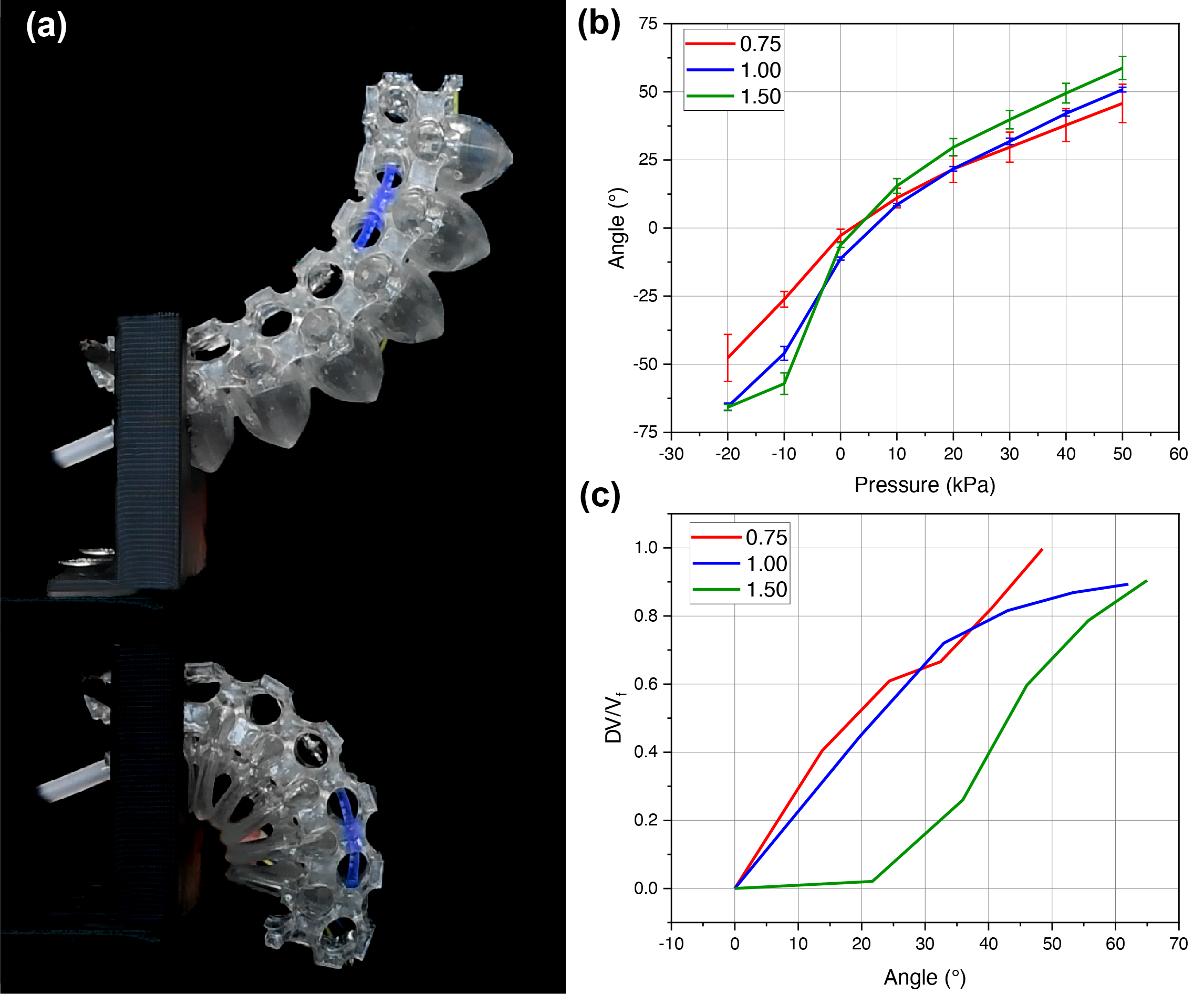}
    \caption{Experimental characterization of the Monolithic Units. 
    (a) Sequential images showing bending of the printed MU under pneumatic actuation. 
    (b) Bending angle as a function of pressure for the three scales ($0.75\times$, $1.0\times$, $1.5\times$). 
    (c) Normalized optical response ($\Delta V/V_f$) versus bending angle, showing reduced sensitivity at small angles for the $1.5\times$ MU due to limited curvature.}
    \label{fig:characterization}
\end{figure}

At smaller scales, the lattice struts become thinner relative to the constant 1.5 mm waveguide thickness, amplifying the local stiffening effect of the sensor, whereas the opposite occurs at the large scale.
The deviation analysis shows that sensors placed distal to the inlet experience lowest deviation from the baseline (optimal solution of $\alpha_4$ for the $0.75\times$ and $1.00\times$ MU), while inclusions closer to the base introduce higher deviations. However, the results of the $1.5\times$ MU show much less deviation across all sensors being placed, with the optimal solution of $\alpha_2$ (Table \ref{tab:sensor_deviation}). 
These same configurations were selected for fabrication and characterization in Figure~\ref{fig:characterization}.

\subsection{Mechanical and Sensor Characterization}

Bending tests were performed on Monolithic Units (MUs) at $0.75\times$, $1.0\times$, and $1.5\times$ scales under pneumatic inputs from $-20$ to $50$ kPa (Figure~\ref{fig:characterization}a).  
All configurations exhibited smooth, repeatable deformation and consistent pressurization–depressurization cycles.  
Measured bending angles increased nonlinearly with pressure, spanning from approximately $-50\text{ to }45^\circ$ for the $0.75\times$ MU, $-65\text{ to }50^\circ$ for $1.0\times$, and $-65\text{ to }55^\circ$ for $1.5\times$ (Figure~\ref{fig:characterization}b). These results indicate that actuation performance is maintained across scales.

Optical responses of the integrated waveguides were characterized concurrently with bending.  
Normalized signal variation ($\Delta V/V_f$) increased monotonically with bending angle (Figure~\ref{fig:characterization}c).  
At small curvatures (0–20$^\circ$), the $1.5\times$ MU showed limited sensitivity due to minimal strain along the embedded waveguide.  
For all scales, the optical output exhibited consistent, reversible behavior over repeated cycles, demonstrating that the embedded sensor reliably tracks deformation without disrupting actuator mechanics.

\subsection{Application Demonstration}
\begin{figure*}[htb!]
    \centering
    \includegraphics[width=\textwidth]{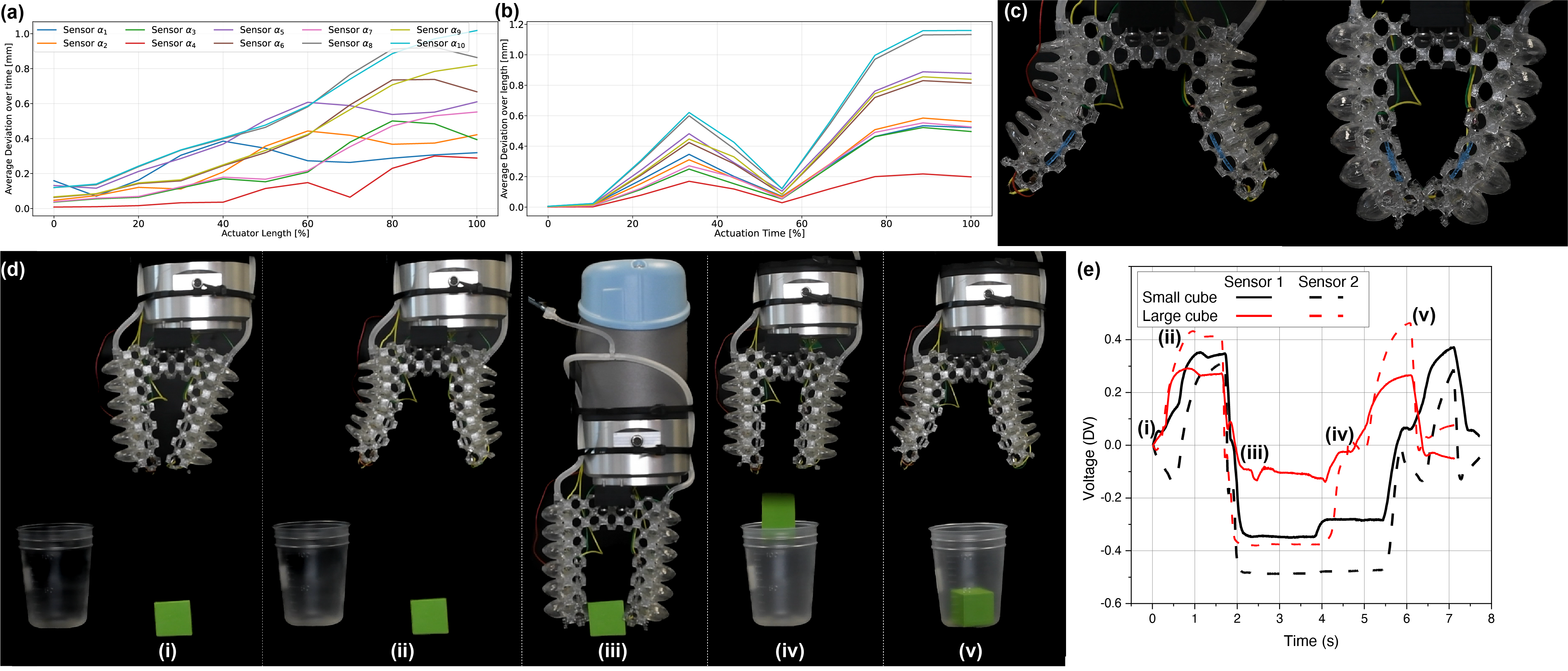}
    \caption{Application of the Monolithic Units (MUs) in a soft gripper. 
    (a) Average nodal deviation along actuator length for all candidate sensors $\alpha_1$–$\alpha_{10}$, showing minimal deviation for most configurations. 
    (b) Average deviation over actuation time, indicating limited mechanical differences between sensor placements. 
    (c) Comparison between baseline and sensorized MU configurations under equivalent actuation inputs. 
    (d) Sequential frames of the MU-based two-finger gripper grasping large (25 mm) cubes. 
    (e) Optical sensor voltage responses during grasp–release cycles for both fingers, demonstrating distinct signal patterns (for both  small (12.5 mm) and large (25 mm) cubes) corresponding to object size and grasp phases (i–v).}
    \label{fig:application}
\end{figure*}

The same ranking of sensor placements from \eqref{eq:objective} is obtained for the gripper design (Table \ref{tab:sensor_deviation}). Two $1.0\times$ MUs configured with the $\alpha_4$ sensor path were combined to form a simple two-finger gripper (Figure~\ref{fig:application}c).  
Each finger retained its nominal actuation behavior and provided independent optical readouts.  
The gripper successfully grasped both a small ($12.5$ mm) and a large ($25$ mm) cube (Figure~\ref{fig:application}d).  
The recorded voltage traces show distinct signal amplitudes for each object size, with higher curvature (small object) producing larger $\Delta V$ (Figure~\ref{fig:application}e).  

\section{Discussion and Future Work}

This study establishes the Monolithic Unit (MU) as a reproducible actuator–lattice–sensor building block for soft robotic systems. The results demonstrate that deterministic design rules and simulation-based evaluation can integrate optical sensing with minimal alteration of actuator behavior. The MU provides a generalizable co-design framework that bridges simulation and fabrication, extending previous monolithic approaches toward quantitatively verifiable integration. Although this effort discretizes \eqref{eq:Jcont} to the designed lattice nodes, the resulting formulation remains a discrete optimization problem, solved here by an exhaustive approach due to the small finite candidate set. 
For larger design spaces, heuristic or gradient-based solvers could replace this approach while preserving the same objective and constraints. 
Future MU designs can therefore extend this framework to perform optimization given a desired bulk stiffness, with the actuator–lattice geometry and corresponding anchor points determined post-optimization.

Future work will focus on refining both material and computational aspects of the framework.  
From a materials perspective, the trade-off between compliance for actuation and stiffness for sensing remains critical; multi-material or graded-lattice strategies may improve this balance.  
On the computational side, the current optimization can be extended into a multi-objective formulation that simultaneously minimizes mechanical deviation and maximizes sensor fidelity or signal sensitivity.  
Such formulations could guide more adaptive design choices, particularly when scaling toward complex systems.

Applications involving dynamically coupled MUs, such as serial chains or undulatory structures, should also be explored.  
In these architectures, local mechanical perturbations may accumulate or amplify, altering global deformation and sensing characteristics.  
Studying these coupled dynamics will be key to translating single-unit behavior to coordinated motion and proprioceptive feedback in multi-segment systems.

Finally, printing orientation and lattice topology warrant further investigation to reduce anisotropy and support-free constraints, enabling more complex geometries without post-processing.  
Together, these directions will extend the MU framework toward fully integrated, multi-functional monolithic systems capable of distributed sensing and adaptive control.

\section{Conclusion}

This work presents a simulation-informed framework for designing and fabricating Monolithic Units (MUs) that integrate pneumatic actuation, lattice reinforcement, and embedded optical sensing within a single printed body. The approach leverages parametrically-defined geometries and numerical evaluation of sensor configurations to maintain mechanical behavior while enabling embedded sensing.  
Demonstrations across multiple scales and a simple gripper application highlight the reproducibility and scalability of the design process.  
Beyond a single actuator, the MU framework provides the first step for systematic co-design of actuation and sensing, serving as a modular platform for constructing larger, interconnected soft robotic systems.  
Future extensions toward multi-objective optimization and dynamic multi-unit architectures will further advance the integration of mechanical and sensing performance in monolithic soft robots.



\bibliographystyle{IEEEtran}

\bibliography{biblio}

\end{document}